\documentclass[letterpaper, 10 pt, journal, twoside]{IEEEtran}

\usepackage{setspace}
\pdfoutput=1 
\usepackage{amsmath,amsfonts}
\usepackage{algorithmic}
\usepackage{algorithm}
\usepackage{array}
\usepackage[caption=false,font=normalsize,labelfont=sf,textfont=sf]{subfig}
\usepackage{textcomp}
\usepackage{stfloats}
\usepackage{url}
\usepackage{verbatim}
\usepackage{graphicx}
\graphicspath{ {figures/} }
\usepackage{cite}
\usepackage{indentfirst}
\usepackage{multirow}
\IEEEoverridecommandlockouts 
\usepackage[margincaption,outercaption,ragged,wide]{sidecap}
\sidecaptionvpos{figure}{t} 
\sidecaptionvpos{table}{t}
\usepackage[amssymb]{SIunits}
\usepackage{booktabs}
\usepackage[colorinlistoftodos,prependcaption,textsize=tiny]{todonotes}
\usepackage{marvosym}
\usepackage{ifsym}
\usepackage{caption}
\usepackage{threeparttable}
\newcommand{\fref}[1]{Fig. \ref{#1}}

\begin{document}

\title{FDCT: Fast Depth Completion for Transparent Objects}

\author{Tianan Li$^{1}$, Zhehan Chen$^{1,\textrm{\Letter}}$,~\IEEEmembership{Member,~IEEE}, Huan Liu$^{2}$,~\IEEEmembership{Member,~IEEE}, and Chen Wang$^{3}$

\thanks{Manuscript received: March, 18, 2023; Revised June, 13, 2023; Accepted July, 17, 2023.}% Use only for final RAL version
\thanks{This paper was recommended for publication by Editor Cesar Cadena Lerma upon evaluation of the Associate Editor and Reviewers' comments.}
\thanks{This work was supported by the National Key R\&D
Program of China: [grant number 2021YFB3401500] and the
Fundamental Research Funds for the Central Universities: [grant number FRF-TP-20-009A3].}%  %Use only for final RAL version</span>}%
\thanks{$^{1}$The authors are with the School of Mechanical Engineering, University of Science and Technology Beijing, Beijing 100083, China. $^{\textrm{\Letter}}$Corresponding author: {\tt chenzh\_ustb@163.com}}% <-this % stops a space}}%
\thanks{$^{2}$Huan Liu is with the Department of Electrical and Computer Engineering, McMaster University, Canada. Email: {\tt liuh127@outlook.com}}%
\thanks{$^{3}$Chen Wang is with the Spatial AI \& Robotics Lab at The Department of Computer Science and Engineering, State University of New York at Buffalo, NY 14260, USA. {\tt\small chenw@sairlab.org}}% <-this % stops a space
\thanks{Digital Object Identifier (DOI): see top of this page.}}

% \markboth{IEEE Robotics and Automation Letters. Preprint Version. Accepted July, 2023}
% {Li \MakeLowercase{\textit{et al.}}: FDCT} 

\markboth{IEEE Robotics and Automation Letters. Preprint Version. Accepted July, 2023}
{Li \MakeLowercase{\textit{et al.}}: FDCT}

\maketitle
% % for arvix
% \thispagestyle{fancy}

% \setlength{\footskip}{5mm}
% \fancyhead{}
% \lhead{}
% \lfoot{\small

% 两倍行距的文字
% \copyright~2023 IEEE.  Personal use of this material is permitted.  Permission from IEEE must be obtained for all other uses, in any current or future media, including reprinting/republishing this material for advertising or promotional purposes, creating new collective works, for resale or redistribution to servers or lists, or reuse of any copyrighted component of this work in other works.

% }
% \cfoot{}
% \rfoot{}

\begin{abstract}
Depth completion is crucial for many robotic tasks such as autonomous driving, 3-D reconstruction, and manipulation.
Despite the significant progress, existing methods remain computationally intensive and often fail to meet the real-time requirements of low-power robotic platforms.
Additionally, most methods are designed for opaque objects and struggle with transparent objects due to the special properties of reflection and refraction.
To address these challenges, we propose a Fast Depth Completion framework for Transparent objects (FDCT), which also benefits downstream tasks like object pose estimation.
To leverage local information and avoid overfitting issues when integrating it with global information, we design a new fusion branch and shortcuts to exploit low-level features and a loss function to suppress overfitting.
This results in an accurate and user-friendly depth rectification framework which can recover dense depth estimation from RGB-D images alone.
Extensive experiments demonstrate that FDCT can run about 70 FPS with a higher accuracy than the state-of-the-art methods.
We also demonstrate that FDCT can improve pose estimation in object grasping tasks.
The source code is available at \url{https://github.com/Nonmy/FDCT}.
\end{abstract}


\IEEEpeerreviewmaketitle

\section{Introduction}

\IEEEPARstart{D}{epth} completion has received increasing attention due to its essential role in various robotic tasks such as autonomous driving \cite{hu2021penet,nazir2022semattnet,yan2022rignet} and manipulation \cite{ichnowski2021dex}.
It aims at predicting per-pixel depth value given a sparse image from a depth sensor such as RGB-D camera \cite{zhang2018deep} or LiDAR \cite{jaritz2018sparse}.
It has been demonstrated that accurate depth completion is able to significantly improve the performance of downstream tasks such as object grasping due to its abundant 3-D information \cite{sajjan2020clear,tang2021depthgrasp,fang2022transcg}.
Despite those progresses, many existing methods still struggle with transparent objects, which are commonly encountered in daily life and modern industry.
% and automatic manipulation needs localization information of the transparent objects. 
% \todo{we need an example here to say transparent object are widely used in graphing}
This is because transparent objects are refractive and reflective, which can cause the optical sensors to produce inaccurate depth measurements \cite{tanaka2016recovering}.
% making it challenging for commercial depth camera to produce accurate depth estimates. 
% However, few methods \cite{gou2021rgb,he2021ffb6d,di2022gpv,li2022diversity} \todo{i have changed "grasping or pose estimation methods" to "depth completion", citation may need to change also} can directly apply to scenarios with mixed transparent and opaque objects, which largely limits their real-world applications.
%In this paper, we will try to solve this problem and propose a fast depth completion framework for transparent objects (FDCT) and show its effectiveness in an object grasping task.
% Therefore, depth completion is essential to the manipulation of transparent objects.
\begin{figure}[!t]
    \centering
    \includegraphics[width=0.9\linewidth]{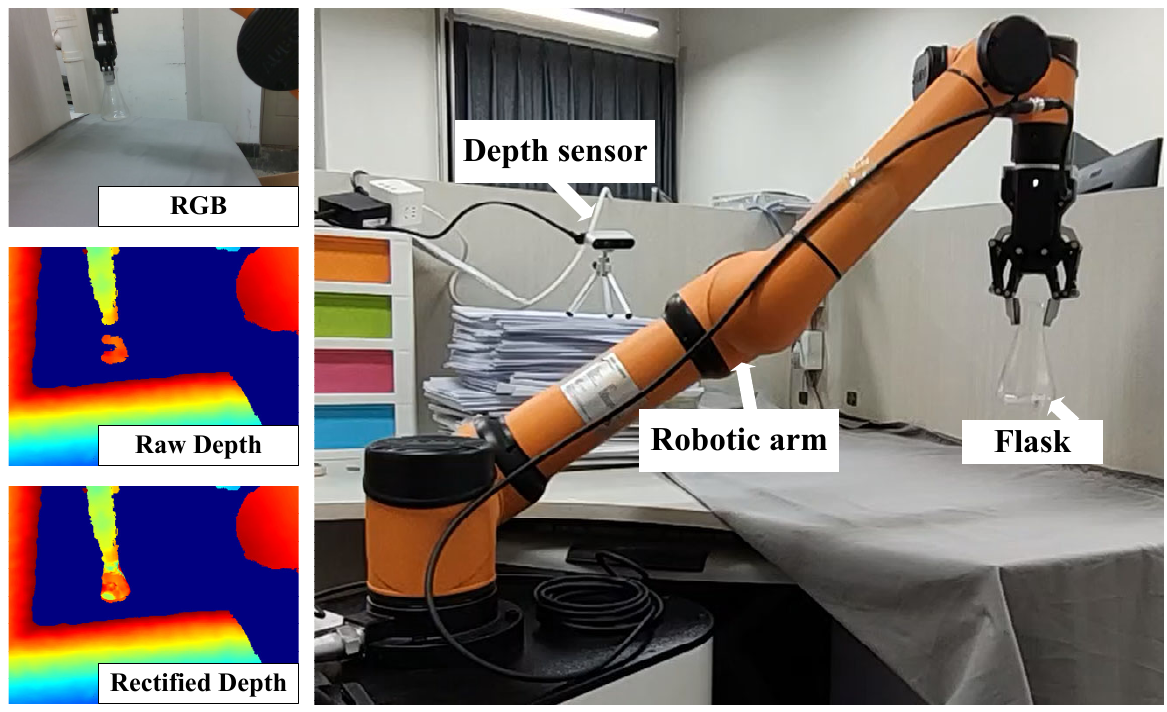}
    \captionsetup{
  font={stretch=0.5}, % 根据需要调整行距的倍数
}
    \caption{\textbf{FDCT} utilizes RGB and raw depth to complete depth estimation in an end-to-end way, which benefits  many robotic tasks including pose estimation and object grasping.}
    \label{fig:figure1}
    \vspace{-0.5cm}
\end{figure}

The field of depth completion for transparent objects is still relatively unexplored, and the current methods either struggle to generalize well or require significant computational resources that may not be suitable for low-power platforms.
% There are only a few depth completion methods focusing on transparent objects but they either cannot generalize well or are quite heavy for low-power platforms.
% attempt to improve the accuracy of pose estimation or grasping by first correcting depth information.
For example, ClearGrasp \cite{sajjan2020clear} uses three networks to predict surface normal, edge, and transparent mask and estimate depth value by global optimization.
It outperforms the monocular depth estimation methods by a large margin, but global optimization is computational heavy and needs well-trained network.
% It outperforms the monocular depth estimation methods by a large margin, and significantly improves the grasping success rate. 
% However, global optimization is time consuming and needs well-trained network.
Zhu et al. \cite{zhu2021rgb} introduces  a method  for local implicit depth prediction and depth refinement based on voxel map, which significantly speeds up the progress. However, the method has limitations in terms of generalization to new environments \cite{fang2022transcg}.
Recently, Fang et al. \cite{fang2022transcg} proposed an encoder-decoder U-Net-based architecture that achieved the state-of-the-art performance in the area.
However, the method has certain limitations, such as directly concatenating the raw depth image to the deep features, makes the network sensitive to depth noise. Furthermore, their network tends to predict depth image with blurry edges.
% However, they straightly concatenates raw depth image to deep features, makes the network sensitive to the noise of the original depth values and their network is slower than \cite{zhu2021rgb}.

% Previous works have problems in inference time and the utilization of raw depth information. Typically, those works use sub-network predicting features such as edge and solve the depth by global optimization or GAN based structure, which is proved to be time consuming. Besides, those low-level features are usually not stable due to sensor noise, lighting conditions and shooting angles, which may lead to extra inaccuracy when they are introduced into the network by independent sub-network. Furthermore, some works contain redundant modules, which disrupts the learning of low-level features and could reduce the speed and accuracy of the network. Other works apply a transparency mask to the original depth map to get a modified one, which destroys the original raw depth information at the outset. 

Low-level feature refers to basic visual elements such as edges, 
 angles etc. As the network deepens, the receptive field increases, and the features transition from local to global, and from low-level to high-level. Research has shown that the low-level feature plays a significant role in depth completion \cite{zuo2016explicit,tao2021dilated,huang2019indoor}.
However, incorporating these features in existing depth completion networks can lead to other issues, such as increased sensitivity to sensor noise and lighting conditions, as well as trade-offs between efficiency and accuracy \cite{sajjan2020clear,fang2022transcg}.
% , such as being too sensitive to the sensor noise, lighting conditions and trade-off between efficiency and accuracy brought about by additional processing \cite{sajjan2020clear,fang2022transcg}. 
This problem becomes more severe when transparent objects need to be handled. 
To address these issues, some methods \cite{zhang2018deep,sajjan2020clear,tang2021depthgrasp} mask out the transparent objects from the original depth images and use a separated sub-network to predict those features.
However, this can result in additional inaccuracies when inferring results. Moreover, using too many local features can lead to overfitting during training, which can negatively impact the generalization ability of the model \cite{ying2019overview}.

To better extract those features while maintaining low computational complexity, we propose a Fast Depth Completion architecture for Transparent objects (FDCT).
Firstly, we streamline the architecture for faster inference by eliminating separated prediction and non-essential modules. 
We adopt a new fusion branch based on shortcut fusion module and cross-layer shortcuts to more efficiently exploit low-level features. 
% Then, a new fusion branch based on Shortcut Fusion Module and Cross-layer Shortcuts are adopted to exploit low-level features more efficiently.
To reduce the semantic differences between raw depth and features, 
we introduce a domain adaptation operation when providing the full original depth image to the network.
% a domain adaption operation is introduced when the full original depth image is provided to the network. 
Meanwhile, we use a downsampling method based on max pooling to accelerate the network while capturing global features. Finally, we design a new loss function to avoid overfitting and fuse local and global features in the decoder. In summary, we make the following contributions:
\begin{enumerate}
  \item We propose a fast architecture for transparent objects depth completion called FDCT. It takes only raw RGB-D images and can assist with downstream tasks, such as object grasping and 3D reconstruction.
  \item To exploit the low-level features, we introduce a fusion branch that utilize Shortcut fusion module and cross-layer shortcuts. We also develop a new loss function to avoid overfitting during the fusion of local and global features, leading to an accurate depth estimation.
%   take advantage of multi-scale and multi-layer information. 
%   Encoder and decoder block are designed to reduce overfitting and may be a reference to future work.
  \item FDCT outperforms the state-of-the-art depth completion methods by at least 16\% while using less parameters and runing at around 70 FPS. Experimental results demonstrate FDCT benefits pose estimation and grasping tasks. 
\end{enumerate}
\section{Related Works}

\subsection{Depth Completion of Transparent Objects}
Depth completion is a long-standing problem in computer vision. Considerable works\cite{eigen2015predicting,chen2016single} infer depth directly from single RGB image. However, these methods can hardly be used for transparent objects. RGB-D-based methods are demonstrated to outperform monocular methods by a large margin in previous work. 
Zhang et al. \cite{zhang2018deep} firstly use a deep learning-based method for depth completion from RGB-D images. They use neural network to predict surface normal and occlusion boundary from RGB image respectively, and then solve the output depths with a global linear optimization regularized by raw depth information. Huang et al.\cite{huang2019indoor} and Tang et al. \cite{tang2021depthgrasp} replace the global optimization part with a neural network, resulting in an end-to-end network.
% , making it an end-to-end network. 

The first work focus on transparent objects depth completion is ClearGrasp\cite{sajjan2020clear}, which employs transparent mask to modify raw depth image. They also constructed the first transparent objects dataset for depth completion and pose estimation. DepthGrasp \cite{tang2021depthgrasp} replaces the global optimization part with adversarial neural network and introduces spectral residual blocks for stability. Both methods are computatioally heavy and require normal information, which is inconvenient to acquire, making them impractical for real-time applications. 
Local Implicit \cite{zhu2021rgb} builds local implicit neural representation on ray-voxel pairs and iteratively refines the depth. It also introduces a large-scale synthetic dataset Omniverse. Their work significantly improves the speed for depth completion. TranspareNet \cite{xu2021seeing} combines depth completion and point cloud completion, using distorted depth to construct point clouds and complete it as a rough representation of depth. 

Recently, Fang et al.\cite{fang2022transcg} proposed a larger-scale real-world dataset TrasCG and a depth completion network for transparent objects (DFNet), it adopts U-Net architecture and uses a lightweight backbone called DenseBlock\cite{huang2017densely}, and achieves SOTA performance with small memory use and model size.

\subsection{Pose Estimation of Transparent Objects}
RGB-D-based methods focusing on opaque rigid objects 6 DoF pose estimation mostly failed to transparent objects due to the inaccuracy of depth value. Early works focusing on transparent objects use traditional features to estimate pose \cite{lysenkov2013recognition,lysenkov2013pose,guo2019transparent}. However, low-level traditional features are not as discriminative as high-level deep features. The RGB-based method KeyPose \cite{liu2020keypose} captures and labels 3D key points from stereo RGB images and estimates pose from these key points. GhostPose* \cite{chang2021ghostpose} further improves this by defining four virtual 3D key points from RGB images and reconstructing 3D coordinates of the key points based on triangulation. Xu et al. \cite{xu20206dof} estimate 6DoF transparent objects pose from a single RGB-D image. Recently, Chen et al. proposed ClearPose \cite{chen2022clearpose}, a large-scale real-world benchmark and dataset for transparent objects, and ProgressLabeller \cite{chen2022progresslabeller}, a semi-automatic method for labeling the 6D pose of transparent objects.

There are works that utilize light field cameras \cite{zhou2019glassloc}, single pixel cameras \cite{mathai20203d}, microscope-camera systems \cite{grammatikopoulou2019three}, and polarization cameras \cite{kalra2020deep} for pose estimation or grasping on transparent objects. However, these methods differ significantly from RGB-D based methods and are less accessible compared to commercial depth cameras like RealSense or Kinect. One potential research direction is the use of neural radiation fields for transparent scenes \cite{ichnowski2021dex,zhou2020lit,mildenhall2021nerf}.
Pose estimation methods using only RGB information will lose multimodal information at the initial stage, and there are only a few methods that can directly estimate the pose of transparent objects from RGB-D information. Thus, we believe that, at the current stage, combining depth completion and specially designed pose estimation methods can bring benefits to the pose estimation of transparent objects.

\section{Method}

\begin{figure*}[thpb]
\centering
\includegraphics[width=0.9\textwidth]{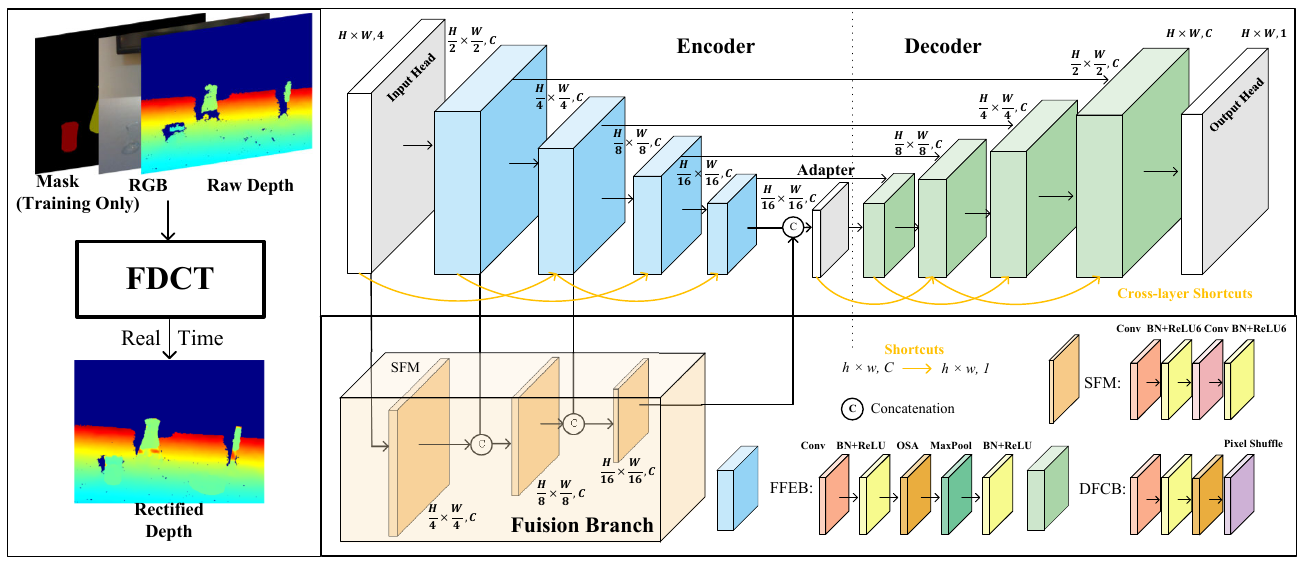}
% \captionsetup{font=small, skip=1pt}
% \captionsetup{
%   font={stretch=0.6}, % 根据需要调整行距的倍数
% }
\caption{The architecture of our proposed depth completion network \textbf{FDCT}. Our proposed method predicts rectified depth from RGB and origin depth images. We build U-Net architecture with Feature Fusion and Extraction Block as encoder (\textbf{FFEB}), and Depth Fusion and Completion Block (\textbf{DFCB}) as decoder. Cross-layer shortcuts and a fusion branch are added to fully use multi-scale and multi-layer information. H,W,C means height, width and channel. Transparent mask is only required for training to compute the loss and raw depth image is provided to each FFEB and DFCB.}
\label{fig:figure2}

\end{figure*}

% \subsection{Overview}
In this section, we present an efficient Depth Completion network for Transparent Objects (FDCT), as illustrated in Figure \ref{fig:figure2}. The network is designed to deliver real-time predictions of rectified depth maps from RGB-D cameras.
% , even on low-power platforms.
We begin by giving an overview of the proposed network, followed by the introduction of the compact design of the encoder and decoder components. Next, we demonstrate the effective utilization of low-level features through the use of our fusion branch and cross-layer shortcut. Finally, we detail the loss functions that are employed during the training of the network.

\vspace{-0.4cm}
\subsection{Overview of Network}
We follow the common practice in \cite{fang2022transcg} to build our network similar to UNet \cite{ronneberger2015u}. The network can predict rectified depth directly from RGB and depth images. The transparent mask is involved only in the training of the network.
 As shown in \fref{fig:figure2}, our network contains an encoder and decoder with skip connection added and a fusion branch serving as a sub-encoder. 
However, the original skip connection in UNet only transmit features from encoder to decoder. As is known,  low-level features contains sufficient position information, which plays an important role in depth completion task \cite{zuo2016explicit,tao2021dilated}. In observing that deeper layer contains less positional information \cite{olah2017feature,peng2021dgfau}, we additionally add shortcuts to every layer (shown as orange connection in Figure \ref{fig:figure2}) to make low-level features available for deeper layers, preserving positional information and leading to more accurate results.  

The encoder of FDCT consists of two parts: the main part is built with Feature Fusion and Extraction Blocks (FFEB), while the subpart is called Fusion Branch and is built with a series of Shortcut Fusion Modules (SFM).  
% SFM integrates features from the encoder layer by layer, preserving the representative features for subsequent decoding and ensuring the integration of multi-layer and multi-scale features. 
The main and sub encoders work together to provide aggregated features to the decoder, preserving low-level information such as pixel position.
Besides, the decoder is constructed with Depth Fusion and Completion Blocks (DFCB).
% , while avoiding harm to high-level information through the use of a new loss function and max pooling-based downsampling method.
In our work, the original depth image is efficiently and accurately utilized by providing it to every FFEB and DFCB through a domain adaptation operation. This is a departure from previous works, which either used a network to predict a transparent mask based on raw depth information \cite{sajjan2020clear,tang2021depthgrasp} (leading to time-consuming two-stage processing) or directly concatenated raw depth to features \cite{fang2022transcg} (leading to sensitivity to noise and inaccuracies).

\begin{figure}[!t]
\centering
\includegraphics[width=0.8\linewidth]{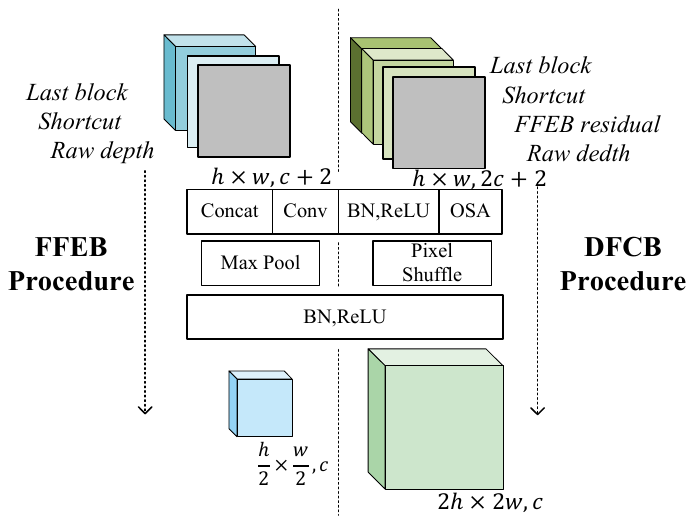}
\caption{The detail of FFEB (left) and DFCB (right).}
\label{fig:figure3}
\vspace{-0.8cm}
\end{figure}

\subsection{Lightweight Encoder and Decoder}
\subsubsection{\bf Feature Fusion and Extraction Block}
% \textcolor{red}{Emphasize the effectiveness and light-weight desgin of our blocks. Give conmparision with dense block.}
% \paragraph{Feature Fusion and Extracting Block}
The backbone plays a crucial role in determining the efficiency of an architecture. Although DenseNet \cite{huang2017densely} has been proven to be effective in previous work \cite{fang2022transcg}, it can be resource-intensive and its shallow layers may not contribute much to the results \cite{lee2019energy}, making it unsuitable for depth completion tasks. To address these issues, we adopt the One-Shot Aggregation module (OSA) \cite{lee2019energy} as the backbone  to effectively extract features.
The Encoder consists of four Feature Fusion and Extraction Blocks (FFEBs), depicted in the left part of Figure \ref{fig:figure3}. Each FFEB takes the original depth map, features from the previous block, and a shortcut from one layer apart as input. The features are concatenated and processed through a data-level fusion. Specifically, we use a convolution operation to force them fusing in a lower dimensional space. This fusion method can reduce the domain gap between deep features and raw depth image \cite{csurka2017comprehensive}. We will further show that it can work better than direct concatenation in Section \ref{sec:analysis}.

% It is a method recommended for raw data fusion\cite{khaleghi2013multisensor} and can reduce the domain gap while fusing them with depth information.

% The features are then concatenated and processed through a convolution operation, which reduces the domain gap while fusing them with depth information. We perform data-level fusion by fusing data in a lower dimensional space. This is a method recommended for raw data\cite{khaleghi2013multisensor}.

% This has been shown to be more effective than concatenating the raw depth directly{\color{blue}\cite{csurka2017comprehensive}}.

Downsampling is performed through max pooling, which not only speeds up the network but also enhances its robustness. To show the effectiveness of the simple max pooling, please refer to our experiments for a thorough analysis of different downsampling methods.
% Backbone is important to the efficiency of the architecture. DenseNet\cite{huang2017densely} is used in the SOTA model DFNet and proved to perform well\cite{fang2022transcg}, however, it is memory and time consuming, and its shallow layers contributes less to the result\cite{lee2019energy}, which might be harmful to depth completion task. Therefore, We use One-Shot Aggregation module (OSA)\cite{lee2019energy} as the backbone of the Encoder for extracting features efficiently. 

% Encoder is formulated by four Feature Fusion and Extracting Blocks (FFEB, the left part in Figure \ref{fig:figure3}). Each FFEB takes the original depth, features from previous block and shortcut from one layer apart as input. Features are concatenated together and subjected to a convolution operation to reduce the domain gap while fusing them with depth information. It is proved to be better than concatenating raw depth directly. 
% Downsampling is achieved by max pooling, which can accelerate network and enable more robust performance. We make comparative analysis of the downsampling methods in ablation study.
% \hspace*{\fill}
\subsubsection{\bf Depth Fusion and Completion Block}
The Decoder is comprised of four Depth Fusion and Completion Blocks (DFCBs), as depicted in the right part of Figure \ref{fig:figure3}. The architecture and the inputs of the DFCBs are similar to those of the FFEBs. A residual connection is established between the corresponding DFCB and FFEB (as indicated by the "FFEB residual" in Figure \ref{fig:figure3}). To restore the resolution, we employ an accurate and efficient technique, pixel shuffle \cite{shi2016real}, which is used as upsampling to reduce computational complexity while taking channel information into consideration.
% Pixel shuffle do upsampling by sub-pixel convolution, reduce computational complexity and take channels into consideration.

% \textcolor{blue}{why efficient and why accurate?}

% Decoder is formulated by four Depth Fusion and Completion Blocks (DFCB, the right part in Figure \ref{fig:figure3}). The architecture and inputs of DFCB are similar to those of FFEB. The residual connection from corresponding DFCB to FFEB is preserved (FFEB residual in Figure \ref{fig:figure3}). We use pixel shuffle\cite{shi2016real}, an efficient and accurate super resolution method, to recover the resolution and channels.

\subsection{Fusion Branch and Cross-layer Shortcuts}
\begin{figure}[!t]
\centering
\includegraphics[width=\linewidth]{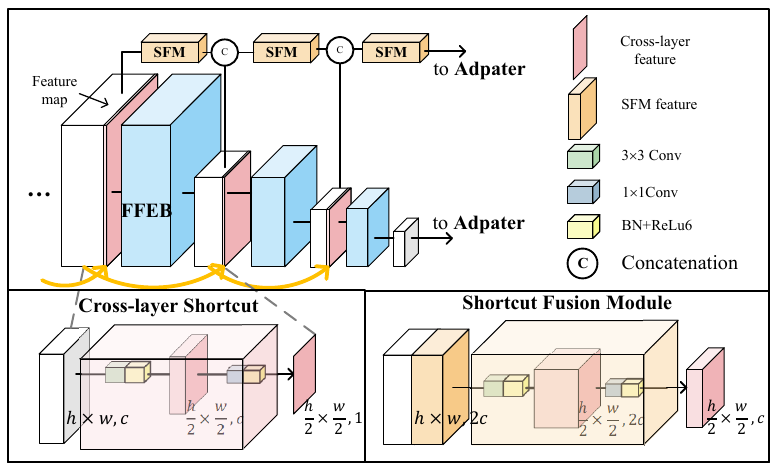}
\caption{The structure of Cross-layer Shortcuts (left) and Shortcut Fusion Module (right).}
\label{fig:figure4}
\vspace{-0.5cm}
\end{figure}

We introduce a novel approach for combining multi-scale and multi-layer features in our model, depicted in Figure \ref{fig:figure4}. Our approach involves the use of Cross-layer Shortcuts and a Fusion Branch, where the Fusion Branch collects the outputs of each encoder layer and integrates them layer by layer using a series of Shortcut Fusion modules (SFMs). Essentially, the Fusion Branch can be thought of as a sub-encoder that contributes to the final output of our model.

% We proposed new Cross-layer shortcut and Fusion Branch to integrate multi-scale and multi-layer features (Figure \ref{fig:figure4}). Fusion branch takes the output of each encoder layer and fuses them layer by layer. It is built by a series of Shortcut Fusion modules (SFM), and can be regarded as a sub-encoder.
%Shortcut Fusion Module (SFM)

% The Fusion Branch is designed to ensure that low-level features remain accessible throughout the decoding process. 
% Many low-level features of the original depth map, such as original pixel values (depth values), edge positioning, etc., are important for depth completion tasks. Adding skip connections allows these features to be transferred to the decoder more directly \cite{hou2021divide}.

% Unlike segmentation, which emphasizes semantic information for class prediction, depth completion focuses on object boundaries and depth values to fill in regions. 
The purpose of the Fusion Branch is to maintain accessibility to low-level features during the coding process. In depth completion tasks, low-level features of the original depth map, such as original pixel values and edge positioning, play a crucial role. 
%The value of some pixels is actually the value that should be predicted, and most of the edge is helpful for rebuild the depth. 
By incorporating skip connections, these features can be directly transferred to the decoder for more effective processing \cite{hou2021divide}.
Therefore, rather than extracting deep semantic information, the Fusion Branch acts like a selector by aggregating features from all levels and exchanging information channel-wise. With this simple structure, it preserves low-level features while fusing multi-level features.

\subsection{Loss Function}
\label{section:Loss}

% In summary, the Fusion Branch and shortcuts provide sufficient low-level features to the network while avoiding overfitting through a carefully designed loss function.
% Fusion branch is designed to keep the low-level features being available to decoding process. Compared to segmentation, where semantic information is important for predicting class, depth completion focus on the boundary of an object and the depth value to fill the region. Instead of deepening digging meaningful semantic information, fusion branch is more like a selector to find the most suitable information by aggregating features from all levels and exchanging information channels-wise. With simple structure, fusion branch preserves the low-level features and fuse multi-level features.

% In general, fusion branch together with shortcuts provide abundant low-level features to the network. But at the same time, too many local features may also lead to overfitting in the training process, affecting the performance of the model. Therefore, we design a loss function to avoid it and experiments show that it is beneficial for depth completion tasks.

% \textcolor{blue}{what is the commonly used loss? (mse+cosine distance of surface normals) What are their problems? Our benefits?}
Too many local features can result in overfitting during the training process, which can negatively impact the performance of the model. To address this, we develop a loss function to prevent overfitting. 
Moreover, having a well-defined optimization objective is critical for deep learning-based methods, while many recent approaches for depth completion neglect the careful design of a proper loss function. 
% Specifically, these methods rely on simple mean squared error (MSE) loss and cosine distance of surface normal (Smooth Loss) as the primary loss functions. 
Specifically, these methods mainly rely on mean squared error (MSE) loss as the primary loss functions. 
This lack of attention to the design of the loss function could potentially lead to suboptimal performance in depth completion tasks.

This observation motivates us to incorporate appropriate losses in model training. Specifically, we note that certain pixels, such as those along edges or affected by sensor noise, often have unstable depth values that would be over penalized by the commonly used mean squared error (MSE) loss. Additionally, using the raw depth image throughout the network with a low-level aware structure can lead to overfitting to these pixels, resulting in poor performance.

To address these issues, we introduce the following loss function that primarily utilizes the Huber Loss, with SSIM loss and smooth loss serving as penalty terms. We demonstrate the necessity of the proposed loss function in the analysis section. All losses are calculated on the transparent regions.
% In the analysis section, we conduct experiments where we modify the weight of these pixels and demonstrate the necessity of our proposed loss function.
% The underlying idea is that some pixels, such as those along edges or affected by sensor noise, often have unstable depth values. Commonly used MSE loss would impose an excessive penalty on these points. Moreover, using the raw depth image throughout the network with a low-level aware structure, overfitting to these pixels can lead to poor performance. To mitigate this issue, we construct a loss function that primarily uses the Huber Loss, with SSIM loss and smooth loss serving as penalty terms. Me make an experiment in the analysis part, modifying the weight of these pixels, and demonstrate The necessity of the proposed loss function.
% The intuition is that, there are usually some pixels whose depth values are unstable e.g., edge, sensor noise. When the raw depth image being used throughout the network with designed low-level awareness structure, overfitting to these pixels could lead to bad performance. To overcome it, we build a loss function with Huber Loss as the main loss, SSIM loss and smooth loss as the penalty terms.

% \noindent\textbf{Huber Loss.}
\hspace*{\fill}
\subsubsection{\bf Huber Loss}

Huber Loss, as described in \eqref{equation:L-huber}, is a well-established loss function in deep learning. Unlike MSE and MAE, Huber Loss is quadratic for small differences below a threshold $\delta$ and linear for larger values, making it less sensitive to outliers in the data. During the training process, $\delta$ is empirically set to a value of 0.1. 
% and will be discussed further in the Limitations section. 
Conceptually, $\delta$ can be viewed as a threshold of error tolerance, where differences below $\delta$ are considered ``acceptable''.
% \begin{equation}
%     L_{\rm Huber} = \left\{ 
%     \begin{matrix}
%         \frac{1}{2}\left( {D_{\rm p} - D_{\rm gt}} \right)^{2} & \left| {D_{\rm p} - D_{\rm gt}} \right| \leq \delta \\
%         \delta\left( {D_{\rm p} - D_{\rm gt}} \right) - \frac{\delta^{2}}{2} & \left| {D_{\rm p} - D_{\rm gt}} \right| > \delta
%     \end{matrix} \right.
%     \label{equation:L-huber}
% \end{equation}
\begin{equation}
    {L_{\rm \rm Huber}} = \begin{cases}
    {\frac{1}{2}( D_{\rm \rm p} - D_{\rm \rm gt})^{2}},&|D_{\rm p} - D_{\rm gt}| \leq \delta, \\
    \delta(D_{\rm p} - D_{\rm gt}) - \frac{\delta^{2}}{2},&|D_{\rm p} - D_{\rm gt}|>\delta,
    \label{equation:L-huber}
\end{cases}
\end{equation}
where, $D_{\rm p}$ and $D_{\rm gt}$ are the predicted and ground-truth depth.

% \noindent\textbf{SSIM Loss.}
\hspace*{\fill}
\vspace{-0.2cm}
\subsubsection{\bf SSIM Loss}
The SSIM (Structural Similarity Index Measure) is a method for assessing the quality of an image, first introduced in \cite{wang2004image}. Instead of computing loss on a pixel-by-pixel basis, SSIM calculates the loss on a global scale, focusing more on the overall structure and distribution of the image, rather than its individual details.
% SSIM (Structural Similarity Index Measure) is an image quality assessment first introduced in 2004\cite{wang2004image}. Rather than computing loss pixel-wise, SSIM computes loss in graph scale. SSIM can draw network's attention more to structure/distribution rather than detail.
\begin{equation}
    L_{\rm SSIM} = \frac{\left( {{2\sigma}_{\rm gt,p} + C_{\rm 2}} \right)\left( {2\mu_{\rm gt}\mu_{\rm p} + C_{\rm 1}} \right)}{\left( \sigma_{\rm gt}^{2} + \sigma_{\rm p}^{2} + C_{\rm 2} \right)\left( \mu_{\rm gt}^{2}{+ \mu}_{\rm p}^{2} + C_{\rm 1} \right)},
    \label{equation:L-ssim}
\end{equation}
where, $\mu$, $\sigma$ are the mean and variance of the depth image, $\sigma_{\rm gt,p}$ is the covariance of ground-truth depth and predicted depth, $C_1$, $C_2$ are constants related to pixel range. 
\hspace*{\fill}
\subsubsection{\bf Smooth Loss}
The smooth loss $L_{\rm smooth}$ is determined by the cosine similarity between the surface normals $S_{\rm pre}$ and $S_{\rm gt}$, which are calculated from the predicted depth and ground-truth depth, respectively. A small value $\epsilon$ is added to prevent division by zero. The smooth loss $L_{\rm smooth}$ models the image edges and is performed at the pixel level \cite{zhu2021rgb,fang2022transcg}: 

% It has been widely adopted by previous research studies \cite{zhu2021rgb,fang2022transcg}:

% $L_{\rm smooth}$ is the cosine similarity between surface normal $S_{\rm pre}$ and $S_{\rm gt}$ computed from predicted depth and ground-truth depth, $\epsilon$ is a small value to avoid division by zero. Smooth loss $L_{\rm smooth}$ also computes at the image scale and is widely used in previous work \cite{zhu2021rgb,fang2022transcg}.
\begin{equation}
L_{\rm smooth} = 1-\frac{\rm S_{pre}\cdot S_{\rm gt}}{\max({||S_{\rm pre}||_{2}\cdot ||S_{\rm gt}||_{2}},\epsilon )}.
\label{equation:L-S}
\end{equation}

Our training loss is described as follow:
\begin{equation}
    L = L_{\rm Huber} + \alpha L_{\rm SSIM} + {\beta L}_{\rm smooth},
    \label{equation:L}
\end{equation}
where $L_{\rm Huber}$ penalizes depth inaccuracy, $L_{\rm SSIM}$ penalizes structural similarity and $L_{\rm smooth}$ penalizes unsmoothness.  $\alpha$ and $\beta$ are the weight parameters.
% Pixels with depth out of range [0.3, 1.5] are regarded as invalid value and do not participate loss calculation.
Pixels with intensity values outside the range of $[0.3, 1.5]$ are considered invalid and are excluded from the loss calculation.

\section{Experiment}

\subsection{Datasets and metrics}

% \noindent\textbf{Dataset.}
\subsubsection{Dataset}
% We adopt three datasets in our experiments, i.e., ClearGrasp \cite{sajjan2020clear}, TransCG \cite{fang2022transcg} and ClearPose \cite{chen2022clearpose}. The ClearGrasp dataset is the pioneering large-scale synthetic dataset that specifically focused on transparent objects. It provids a large-scale synthetic dataset as well as a real-world benchmark. The TransCG dataset comprises 57K RGB-D images from 130 different real-world scenes. 
% ClearPose dataset contains 350K RGB-D images of 63 household objects in real-world settings. Depth completion experiments and generalization verification (reported respectively in Section \ref{sec:depth} and \ref{sec:generalization}) are conducted on ClearGrasp, TransCG and ClearPose. Ablation study (reported in Section \ref{sec:ablation}) is performed on TransCG.
We use three datasets including ClearGrasp \cite{sajjan2020clear}, TransCG \cite{fang2022transcg}, and ClearPose \cite{chen2022clearpose}. The ClearGrasp dataset is a pioneering large-scale synthetic dataset that specifically focuses on transparent objects. It provides a large-scale synthetic dataset as well as a real-world benchmark. The TransCG dataset comprises 57K RGB-D images from 130 different real-world scenes. The ClearPose dataset contains 350K RGB-D images of 63 household objects in real-world settings. 
\begin{table}[!t]
\renewcommand{\arraystretch}{1.05}
\setlength{\tabcolsep}{5pt}
\caption{Ablation study. We show the impact of progressively substituting the components of the DFNet with ours. \label{tab:table1}
}
\centering
\resizebox{\linewidth}{!}{%
\begin{threeparttable}
\begin{tabular}{cccccccccc}
\toprule
Model   & RMSE  & REL   & MAE   & $\delta$1.05 & $\delta$1.10 & $\delta$1.25          & Time(s)& Para(M) & Size (MB)   \\ \midrule
DFNet\cite{fang2022transcg}          & 0.018 & 0.027 & 0.012 & 83.76 & 95.67 & 99.71          & 0.0244        & 1.25 & 4.819 \\ \midrule
Huber Loss &0.017   &0.027  &0.012  &84.10  &95.82  &99.74 &0.0244  &1.25   &4.819  \\ \midrule
New Loss        & 0.017 & 0.026 & 0.012 & 84.42 & 96.30 & \textbf{99.81} & 0.0244        & 1.25 & 4.819 \\ \midrule
SF* & 0.017 & 0.024 & 0.011 & 86.18 & 96.67 & 99.79          & 0.0218        & 1.02 & 3.919 \\ \midrule
Ours(s)* & 0.016          & 0.024          & 0.011          & 86.22          & 96.64          & \textbf{99.81} & \textbf{0.0143} & \textbf{0.39} & \textbf{1.518} \\ \midrule
Ours       & \textbf{0.015} & \textbf{0.022} & \textbf{0.010} & \textbf{88.18} & \textbf{97.15} & \textbf{99.81} & 0.0153          & 1.25          & 4.803          \\
\bottomrule
\end{tabular}%
% \multicolumn{10}{l}{Note: NL* represents New Loss, SF* represents Shortcut Fusion and Ours(s)* represents Ours(slim).}
\begin{tablenotes}
\footnotesize
\item Note: SF* represents Shortcut Fusion and Ours(s)* represents Ours(slim).
\end{tablenotes}

\end{threeparttable}
}

\end{table}
% \vspace{-0.5cm}
\subsubsection{Metrics}
For evaluating the performance of our depth completion model, we employ four common metrics: RMSE, REL, MAE and Threshold $\delta$ (where $\delta$ is set to 1.05, 1.10, and 1.25). These metrics are calculated only on the transparent areas, as determined by transparent masks.
% Me use common metrics RMSE, REL, MAE and Threshold $\delta$ ($\delta$ is set to 1.05, 1.10 and 1.25) to evaluate our model. All metrics are calculated on the transparent areas according to transparent masks.

% We use three metrics to evaluate performance on pose estimation task. The average closest point distance (ADD-S)\cite{xiang2017posecnn} calculates the mean distance from each 3D model point to its closest neighbor on the target model. Followed DenseFusion\cite{wang2019densefusion} we report the area under the ADD-S curve (AUC) and the percentage of ADD-S smaller than 2cm ($<$2cm).

\subsection{Implementation Details}
% \noindent
% \textbf{Network configuration.}
\subsubsection{\bf Network Configuration}
% \textcolor{blue}{
In the network architecture, the number of hidden channels, \textbf{$C$}, is set to 64. Each FFEB/DFCB contains a single OSA module. Each OSA module is composed of 5 layers with stage channels of 20. The SFM module maintains \textbf{$C$} channels throughout the pipeline, while cross-layer shortcuts have only 1 channel. Residual connections between the encoder and decoder retain only \textbf{$C$} channels. The input head module and output head module use $3\times3$ convolution to adjust the number of channels and resolution (with resolution changes only occurring in the input head module). For the slim version, \textbf{$C$} is set to 32, and the OSA block contains 4 layers with stage channels of 16.
% }
% The hidden channels \textbf{$C$} in the network is set to 64. Each FFEB/DFCB contains one OSA module, in which, we use 5 layers per block and set stage channels \textbf{$C'$} to 20. SFM keeps \textbf{$C$} channels throughout the pipeline while cross-layer shortcuts take 1 channel only. Residual connections between encoder and decoder just keep channel \textbf{$C$}. $3\times3$ convolution is used in the input head module and the output head module to modify channels and resolution (resolution modified in the input head module only). For slim version, \textbf{$C$} is set to 32, \textbf{$C'$} is set to 16 and uses 4 layers per OSA block.

\subsubsection{\bf Training Details}
% \noindent
% \textbf{Training details.}
All experiments are carried out using the AdamW optimizer with an initial learning rate of $10^{-3}$. The learning rate is reduced by half after 5, 15, 25, and 35 epochs, and training continues for a total of 40 epochs with a batch size of 32. The threshold $\delta$ is kept constant at 0.1 during the training process. The weights $\alpha$ and $\beta$ for the loss function are set to 0.1 and 0.001, respectively. The images are resized to $320\times240$ for both training and testing. The experiments were conducted using an NVIDIA GeForce RTX 3090 GPU.
% We use AdamW optimizer with initial learning rate of $10^{-3}$ and multi-step learning rate scheduler which decays the learning rate by half after 5, 15, 25, 35 epochs. We train the model for 40 epochs with the batch size of 32. Threshold $\delta$ keeps 0.1 during training. Considering loss, we set $\alpha=0.1$, $\beta=0.001$. For all methods, we scale the images to $320\times240$ during training and testing. We use NVIDIA GeForce RTX 3090 for training and testing. 

 % Depth completion task and generalization ability are tested on ClearGrasp, TransCG and ClearPose. Pose estimation task is carried out on the set1 of ClearPose, since Clearpose has an accurate pose annotation without sticker. We use typical network DenseFusion\cite{wang2019densefusion} as pose estimation network. Following the learning strategy of DenseFusion, we train the network on 12G NVIDIA TITAN Xp GPU for 5 epochs with batch size of 128. The margin of refinement is set to 0.03. For fair comparison, we evaluate others works using their released source codes and optimal hyper-parameters or statistics reported in their paper.

\subsection{Ablation study} \label{sec:ablation}
We conduct an ablation study to investigate the effectiveness of our proposed components, including new loss function, fusion branch, cross-layer shortcut and backbone structure. We take DFNet as the baseline method since it is constructed following the UNet structure. We gradually replace its original components with our proposed ones and show the influence of using our proposed components. All the experiments of the ablation study are conducted on the TransCG dataset.

% In view that DFNet is also constructed based on UNet, We here gradually replace its original components by our proposed. This study is conducted on TransCG dataset.
% To study the impact of each component in our proposed method, we perform experiments with different configurations of loss functions, network architecture, and backbones. Our method is compared against the recent transparent object depth completion work DFNet, which serves as our baseline. The ablation study experiments are all performed on the TransCG dataset.
% To verify the effectiveness of each component in our method, we evaluate the performance w.r.t. different configurations of loss functions, network architecture, and backbones. We use recently proposed transparent objects depth completion work DFNet as baseline. Ablation study is carried out on TransCG.

\subsubsection{\bf Loss Function}
The training of DFNet employs the mean squared error (MSE) and smooth loss as its loss function. However, these simple loss functions can lead to overfitting to local features, which makes the model more sensitive to the noise from low-level features such as edges and positions, negatively impacting its accuracy. To validate our proposed loss function, we first replace the MSE loss with Huber loss in DFNet and term it Huber Loss. 
And then, we replaced the loss function of DFNet with ours (New Loss in Table \ref{tab:table1}), leaving all other aspects unchanged. By comparing the firt two row in Table \ref{tab:table1}, we observe an imporovement of using Huber loss. It can be further observed by comparing New Loss with DFNet that all metrics showed improvement without requiring any additional parameters. 

\subsubsection{\bf Fusion Branch and Cross-layer Shortcuts}
In order to evaluate the impact of our proposed fusion branch and cross-layer shortcuts, we make changes to DFNet's architecture. First, we remove the redundant CDC blocks in DFNet from its skip connections, in line with our insight of preserving low-level features and the purpose of lightweight design. Then, we added cross-layer shortcuts and a fusion branch to the modified network. It can be seen in Table \ref{tab:table1} that adopting this new architecture (referred to as Shortcut Fusion), almost all metrics show improvement with fewer parameters. 

\subsubsection{\bf Backbone}
We finally replace the denseblock in DFNet with our OSA module and utilized max pooling as the downsampling method. This final modification has transformed DFNet into our network. As shown in Table \ref{tab:table1}, our network outperforms the previous state-of-the-art (SOTA) by at least 16\% on difference-based metrics and improves ratio-based metrics by up to 4.42\%, resulting in a new SOTA performance. To make it practical for low-power robots, we created a slim version to balance speed and accuracy.

\subsection{Depth Completion Experiments} \label{sec:depth}

We compare our method with others on synthetic dataset ClearGrasp and real-world dataset TransCG. The quantitative results are respectively reported in Table \ref{tab:table2} and Table \ref{tab:table3}. Our proposed network surpasses others in almost every metric on these datasets which contain  synthetic and real-world scenes. Our method achieves a new state-of-the-art performance with a smaller model size and faster inference time, making it a highly competitive solution in this field.
%except on ClearGrasp synthetic validation set. It may be result of that the local implicit depth function which is environment-dependent, as well as the extra training data. 

\begin{figure}
    \centering
    \includegraphics[width=0.9\linewidth]{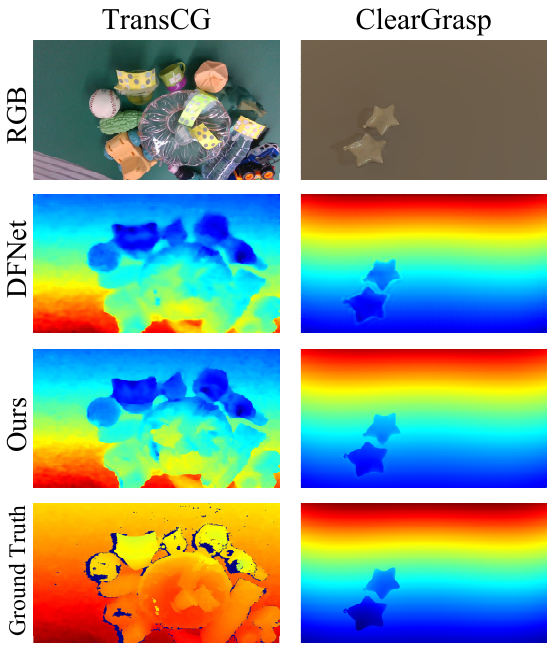}
    \caption{Depth completion examples on TransCG and ClearGrasp dataset.}
    \label{fig:figure5}
% \vspace{-0.5cm}
\end{figure}

% {\color{blue}
Specifically, our method outperforms the other methods by a larger margin in terms of REL and $\delta1.05$ metrics. This indicates its robustness to noise in the raw depth information, as these metrics are computed based on relative values and are sensitive to noise. Additionally, the gap between our method and others is larger in tests involving novel objects in ClearGrasp (CG Syn-novel in Table \ref{tab:table4} and the ClearGrasp column in Figure \ref{fig:figure5}), indicating that our method has a better ability to generalize to unseen objects. The qualitative results is reported in Figure \ref{fig:figure5}. The prediction of our method exhibits a clearer boundary and finer details than DFNet.
% }
% Specifically, our method has a bigger gap in REL and $\delta1.05$ to others most of the time. It demonstrates that our method is more stable to the noise in raw depth information of pixels, because these metrics are computed by relative value and significantly affected by noise. Noteworthy, the gap between our method and others getting bigger in the test of novel objects in most cases, indicates our method is able to generalize better to unseen objects.

\begin{table}[!t]
\caption{Depth Completion Result on TransCG dataset.}
\label{tab:table2}

\centering
\resizebox{\linewidth}{!}{%
\begin{tabular}{ccccccccc}
\toprule
Model & RMSE  & REL   & MAE   & $\delta1.05$ & $\delta1.10$ & $\delta1.25$ & Time ($\second$)   & Size ($\mega$B)    \\ \midrule
ClearGrasp\cite{sajjan2020clear}   & 0.054 & 0.083 & 0.037 & 50.48 & 68.68 & 95.28 & 2.281          & 934          \\
LIDF-Refine\cite{zhou2021pr}  & 0.019 & 0.034 & 0.015 & 78.22 & 94.26 & 99.80 & 0.018          & 251          \\
DFNet\cite{fang2022transcg}        & 0.018 & 0.027 & 0.012 & 83.76 & 95.67 & 99.71 & 0.024          & 4.8          \\
Ours (slim)   & 0.017 & 0.025 & 0.011 & 85.53 & 96.46 & 99.79 & \textbf{0.014} & \textbf{1.6} \\
Ours & \textbf{0.015} & \textbf{0.022} & \textbf{0.010} & \textbf{88.18} & \textbf{97.15} & \textbf{99.81} & 0.015 & 4.8 \\ \bottomrule
\end{tabular}}
% \vspace{-0.5cm}
\end{table}

\begin{table}[!t]
\renewcommand{\arraystretch}{0.9}
\caption{Depth Completion Results on ClearGrasp dataset\label{tab:table3}}
\centering
\resizebox{\linewidth}{!}{%
\begin{tabular}{ccccccc}
\toprule
\multicolumn{1}{c}{Model/Metric} &
  \multicolumn{1}{c}{RMSE} &
  \multicolumn{1}{c}{REL} &
  \multicolumn{1}{c}{MAE} &
  \multicolumn{1}{c}{$\delta$1.05} &
  \multicolumn{1}{c}{$\delta$1.10} &
  $\delta$1.25 \\ \midrule
\multicolumn{7}{c}{Train CG Test CG Syn-novel} \\ \midrule
\multicolumn{1}{c}{ClearGrasp} &
  \multicolumn{1}{c}{0.040} &
  \multicolumn{1}{c}{0.071} &
  \multicolumn{1}{c}{0.035} &
  \multicolumn{1}{c}{42.95} &
  \multicolumn{1}{c}{80.04} &
  98.10 \\ 
\multicolumn{1}{c}{Local Implicit} &
  \multicolumn{1}{c}{\underline{0.028}} &
  \multicolumn{1}{c}{\underline{0.045}} &
  \multicolumn{1}{c}{\underline{0.023}} &
  \multicolumn{1}{c}{\underline{68.62}} &
  \multicolumn{1}{c}{\underline{89.10}} &
  \underline{99.20} \\ 
\multicolumn{1}{c}{DFNet} &
  \multicolumn{1}{c}{0.032} &
  \multicolumn{1}{c}{0.051} &
  \multicolumn{1}{c}{0.027} &
  \multicolumn{1}{c}{62.59} &
  \multicolumn{1}{c}{84.37} &
  98.39 \\ 
\multicolumn{1}{c}{FDCT (Ours)} &
  \multicolumn{1}{c}{\textbf{0.025}} &
  \multicolumn{1}{c}{\textbf{0.040}} &
  \multicolumn{1}{c}{\textbf{0.021}} &
  \multicolumn{1}{c}{\textbf{71.66}} &
  \multicolumn{1}{c}{\textbf{92.95}} &
  \textbf{99.64} \\ \midrule
\multicolumn{7}{c}{Train CG Test CG Syn-known} \\ \midrule
\multicolumn{1}{c}{Local Implicit} &
  \multicolumn{1}{c}{\textbf{0.012}} &
  \multicolumn{1}{c}{\textbf{0.017}} &
  \multicolumn{1}{c}{\textbf{0.009}} &
  \multicolumn{1}{c}{\textbf{94.79}} &
  \multicolumn{1}{c}{\textbf{98.52}} &
  99.67 \\ 
\multicolumn{1}{c}{ClearGrasp} &
  \multicolumn{1}{c}{0.044} &
  \multicolumn{1}{c}{0.047} &
  \multicolumn{1}{c}{0.033} &
  \multicolumn{1}{c}{71.23} &
  \multicolumn{1}{c}{92.60} &
  98.24 \\ 
\multicolumn{1}{c}{DFNet} &
  \multicolumn{1}{c}{0.018} &
  \multicolumn{1}{c}{0.023} &
  \multicolumn{1}{c}{0.013} &
  \multicolumn{1}{c}{88.85} &
  \multicolumn{1}{c}{97.57} &
  \underline{99.92} \\ 
\multicolumn{1}{c}{FDCT (Ours)} &
  \multicolumn{1}{c}{\underline{0.015}} &
  \multicolumn{1}{c}{\underline{0.020}} &
  \multicolumn{1}{c}{\underline{0.012}} &
  \multicolumn{1}{c}{\underline{90.53}} &
  \multicolumn{1}{c}{\underline{98.21}} &
  \textbf{99.99} \\ \bottomrule

\end{tabular}%
% \tablen}
}
\end{table}

\subsection{Generalization Experiment} \label{sec:generalization}
% The generalization capability of a network is essential for practical applications. We evaluated the generalization ability of our proposed method from two perspectives: from synthetic images to real-world images and from one real-world dataset to another. The results of our experiments, shown in Table \ref{tab:table6}, indicate that our method (FDCT) has a comparable generalization capability to the state-of-the-art methods in cross-dataset evaluations, and it outperforms similar works in the synthetic-to-real test. However, it lags behind methods that focus solely on sim-to-real (noted as "local implicit*").
% The generalization ability of a network is critical for real-world application. The proposed method has a generalization ability that can be trained on synthetic data and aply to real world scene (syn-to-real) or trained on one real world dataset TransCG and adap to ClearGrasp (real-to-real). Comparison result is reported in Table \ref{tab:table4}. It shows that although there is still a certain gap compared with the method Local Implicit designed for syn-to-real; compared with the similar method DFNet, our method achieves a better result in the syn-to-real setting, and a competitive result in the syn-to-syn setting.
The generalization ability of a network is critical for real-world application. Our proposed method exhibits a high degree of generalization, being able to be trained on synthetic data and applied to real-world scenes (syn-to-real), or trained on one real-world dataset TransCG and adapted to the other real-world dataset (real-to-real), such as ClearGrasp. Comparison results are reported in Table \ref{tab:table4}, which show that while there is still a certain gap compared to the syn-to-real method (Local Implicit \cite{zhu2021rgb}), our method achieves better results in the syn-to-real setting when compared to the similar method DFNet, and competitive results in the real-to-real setting.

% We inspect the generalization ability of our proposed method from two aspects, from synthetic image to real-world image and from one real-world dataset to another. Experiment results in Table \ref{tab:table5} show that FDCT has a similar generalization ability to previous SOTA in cross-dataset and get better result in synthetic-to-real test compared to similar work, but is far below to methods focusing on sim-to-real.

% Since both datasets comprise real-world image, we train models on TransCG and test it on ClearGrasp real-world set for cross-dataset test. DFNet outperformed other method with a huge gap in generalization test and is chosen to be compared with ours. Comparison result is reported in Table \ref{tab:table5}. Our method outperforms the closest work in all metrics both for known and novel objects in synthetic-to-real test. There is a bigger gap between DFNet and ours in terms of novel objects. It might owe to a better utilization of RGB cues. Our method gets similar results to DFNet in cross dataset test, showing that our method has the ability to generalize from real-world dataset to another. With a series of real-world transparent objects datasets being proposed, we believe that the generalization ability in real-world is more important than sim-to-real.

% {\color{blue}

\begin{table}[!t]
\caption{
% Result of Synthetic to Real and Cross Dataset Generalization Experiment
Generalization test on syn-to-real and real-to-real.}
\label{tab:table4}
\renewcommand{\arraystretch}{0.90}
\centering
\resizebox{\linewidth}{!}{%
% \begin{threeparttable}
\begin{tabular}{ccclclclclcl}
\toprule
\multicolumn{1}{c}{Model/Metric} &
  \multicolumn{1}{c}{RMSE} &
  \multicolumn{2}{c}{REL} &
  \multicolumn{2}{c}{MAE} &
  \multicolumn{2}{c}{$\delta$1.05} &
  \multicolumn{2}{c}{$\delta$1.10} &
  \multicolumn{2}{c}{$\delta$1.25} \\ \midrule
\multicolumn{12}{c}{Train CG Test CG Real-known (syn-to-real)} \\ \midrule
\multicolumn{1}{c}{Local Implicit\cite{zhu2021rgb}} &
  \multicolumn{1}{c}{\textbf{0.028}} &
  \multicolumn{2}{c}{\textbf{0.033}} &
  \multicolumn{2}{c}{\textbf{0.020}} &
  \multicolumn{2}{c}{\textbf{82.37}} &
  \multicolumn{2}{c}{\textbf{92.98}} &
  \multicolumn{2}{c}{\textbf{98.63}} \\ 
\multicolumn{1}{c}{DFNet} &
  \multicolumn{1}{c}{0.068} &
  \multicolumn{2}{c}{0.107} &
  \multicolumn{2}{c}{0.059} &
  \multicolumn{2}{c}{32.42} &
  \multicolumn{2}{c}{56.88} &
  \multicolumn{2}{c}{91.47} \\ 
\multicolumn{1}{c}{FDCT (Ours)} &
  \multicolumn{1}{c}{\underline{0.065}} &
  \multicolumn{2}{c}{\underline{0.103}} &
  \multicolumn{2}{c}{\underline{0.057}} &
  \multicolumn{2}{c}{\underline{33.08}} &
  \multicolumn{2}{c}{\underline{59.81}} &
  \multicolumn{2}{c}{\underline{91.70}} \\ \midrule
\multicolumn{12}{c}{Train CG Test CG Real-novel (syn-to-real)} \\ \midrule
\multicolumn{1}{c}{Local Implicit\cite{zhu2021rgb}} &
  \multicolumn{1}{c}{\textbf{0.025}} &
  \multicolumn{2}{c}{\textbf{0.036}} &
  \multicolumn{2}{c}{\textbf{0.020}} &
  \multicolumn{2}{c}{\textbf{76.21}} &
  \multicolumn{2}{c}{\textbf{94.01}} &
  \multicolumn{2}{c}{\textbf{99.35}} \\ 
\multicolumn{1}{c}{DFNet} &
  \multicolumn{1}{c}{0.051} &
  \multicolumn{2}{c}{0.088} &
  \multicolumn{2}{c}{0.046} &
  \multicolumn{2}{c}{31.23} &
  \multicolumn{2}{c}{64.66} &
  \multicolumn{2}{c}{97.77} \\ 
\multicolumn{1}{c}{FDCT (Ours)} &
  \multicolumn{1}{c}{\underline{0.043}} &
  \multicolumn{2}{c}{\underline{0.073}} &
  \multicolumn{2}{c}{\underline{0.038}} &
  \multicolumn{2}{c}{\underline{39.42}} &
  \multicolumn{2}{c}{\underline{75.54}} &
  \multicolumn{2}{c}{\underline{99.09}} \\ \midrule
\multicolumn{12}{c}{Train TCG Test CG Real-novel (real-to-real)} \\ \midrule
\multicolumn{1}{c}{Local Implicit\cite{zhu2021rgb}} &
  \multicolumn{1}{c}{0.152} &
  \multicolumn{2}{c}{0.225} &
  \multicolumn{2}{c}{0.139} &
  \multicolumn{2}{c}{9.86} &
  \multicolumn{2}{c}{20.63} &
  \multicolumn{2}{c}{46.02} \\ 
\multicolumn{1}{c}{DFNet} &
  \multicolumn{1}{c}{\textbf{0.041}} &
  \multicolumn{2}{c}{\textbf{0.054}} &
  \multicolumn{2}{c}{\textbf{0.031}} &
  \multicolumn{2}{c}{\textbf{62.74}} &
  \multicolumn{2}{c}{\textbf{83.31}} &
  \multicolumn{2}{c}{\textbf{97.33}} \\ 
\multicolumn{1}{c}{FDCT (Ours)} &
  \multicolumn{1}{c}{\textbf{0.041}} &
  \multicolumn{2}{c}{\underline{0.055}} &
  \multicolumn{2}{c}{\underline{0.032}} &
  \multicolumn{2}{c}{\underline{61.23}} &
  \multicolumn{2}{c}{\underline{82.84}} &
  \multicolumn{2}{c}{\underline{97.28}} \\ \bottomrule
\end{tabular}
%     \begin{tablenote}
%         \footnotesize
%         \item [*]Local Implicit is method aiming at sim-to-real.
%     \end{tablenote}
% \end{threeparttable}
}
%\vspace{-0.5cm}
\end{table}
\begin{figure}
    \centering
    \includegraphics[width=1.0\linewidth]{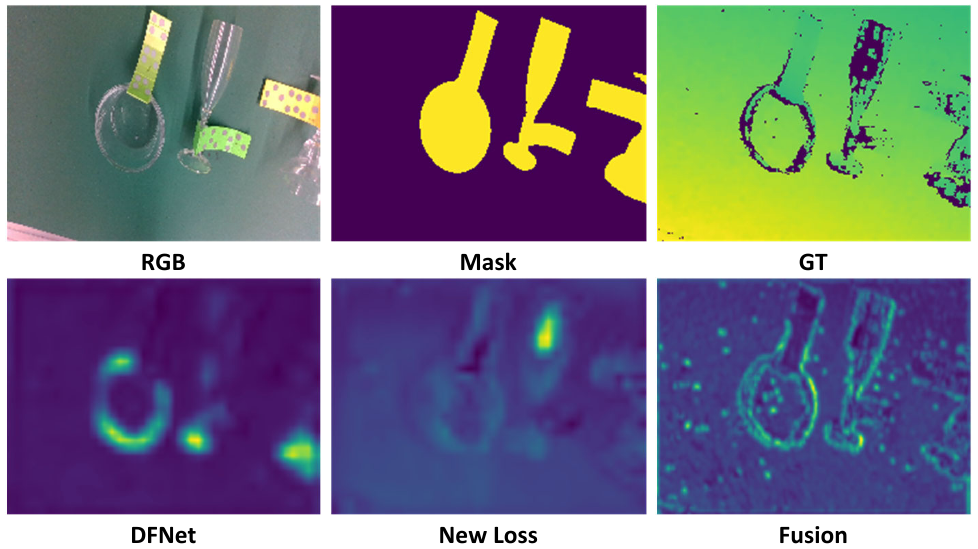}
    \caption{
    % Visualization of the feature maps of the last encoder in FDCT. (a) is a image from TransCG and (b) is a image from ClearGrasp.
    Visualization of the feature maps of the second encoder. 
    \vspace{-1.0cm}
    }
    \label{fig:figure6}
\end{figure}
\subsection{Analysis} \label{sec:analysis}
In our proposed method, the loss function plays a crucial role in enabling the network to focus on structural information and alleviate the effects of unstable pixels. However, this focus on structural information may come at the expense of some details. On the other hand, the fusion branch and shortcuts draw attention to the details, which can introduce extra redundancy. Nonetheless, the use of maxpooling facilitates lossy and aggressive downsampling, which can reduce redundancy and improve robustness. The convolution based fusion method makes better use of the raw depth image. All components work together and complement each other to achieve the best possible balance between structural information and details. In this section, we analyze the four critical components of our method and demonstrate their effectiveness.

\subsubsection{Influence of loss term}
% As we mentioned above, some unstable pixels can unwantedly make big penalty to the loss. By computing the gradient of the depth image and applying Gaussian blur, we manually created a feature to represent these pixels. As the weights of these pixels were reduced, the model's performance improved (as seen in Experiment of weight in Table \ref{tab:table5}), indicating the importance of treating pixels differently and pointing out the necessity of the so designed loss function. However, the side effect of such loss function is that the network pays too much attention to the structure and ignores some details. The highlighted area of the feature map changes from dotted to regional in the Loss column in Figure \ref{fig:figure6}.
As mentioned in \ref{section:Loss}, unstable pixels can have a significant negative influence on the calculation of the training loss. To illustrate this issue, we manually created a feature to represent these pixels by computing the gradient of the depth image and applying a Gaussian blur. By reducing the weights of these pixels, we observed an improvement in the model's performance (as seen in the experiment of weight in Table \ref{tab:table5}), highlighting the importance of treating pixels differently and emphasizing the necessity of the used loss functions (especially the Huber Loss). Qualitatively, as shown in Figure \ref{fig:figure6}, the New Loss model places greater emphasis on the overall structure of transparent objects, while DFNet primarily focuses on local information. The downside of such a loss function is that the network may ignore some details.
\begin{figure}[!t]
\centering
\includegraphics[width=0.9\linewidth]{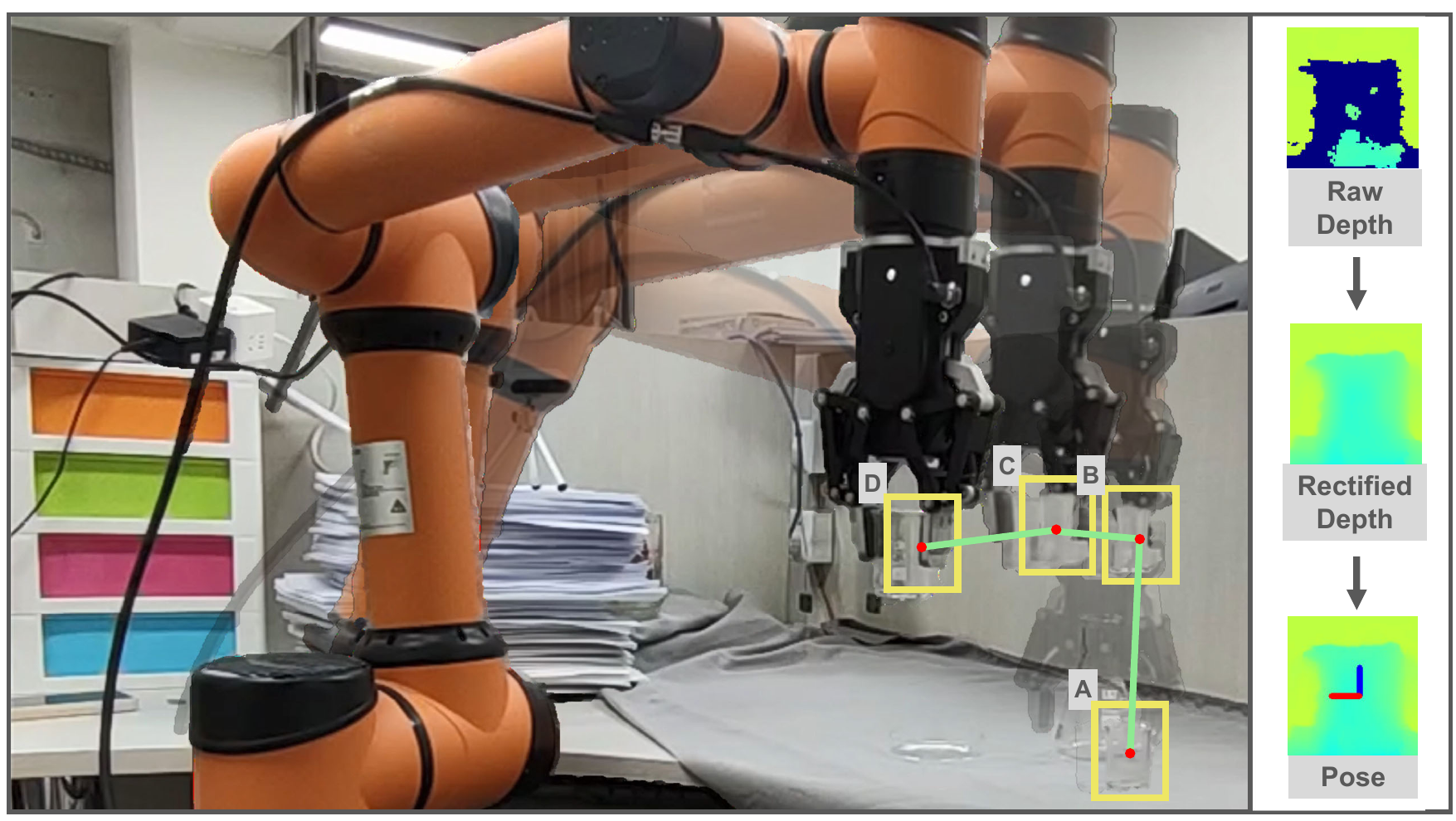}
\captionsetup{font=small, skip=1pt}
\caption{Live demonstration of robot grasping. Dark blue pixels in raw depth indicate the missing depth value.}
\label{fig:figure7}
\vspace{-0.5cm}
\end{figure}

\subsubsection{Low-level feature preservation}
% Fusion branch and cross-layer shortcuts alleviate the indistinct boundaries and perceptual details by taking more low-level cues into consideration. The highlighted area of the feature map changes from regional to scattered in the Fusion column in Figure \ref{fig:figure6}. Loss function and low-level feature awareness components together make a good trade-off between detail and structure information.
The fusion branch and cross-layer shortcuts help alleviate the issue of blurry boundaries and low perceptual details by incorporating more low-level cues. As a result, more low-level features such as object edges and holes are preserved in the feature map of Fusion model in Figure \ref{fig:figure6}. The combination of the loss function and low-level feature awareness components strikes a good balance between detail and structural information.

\subsubsection{Influence of downsampling}
Our hypothesis is that the use of max pooling as a lossy downsampling method can mitigate the side effects of the low-level awareness components while reducing the number of parameters. The results in Table \ref{tab:table5} that are noted as ``Experiment of downsampling'' support our viewpoint. It can be observed that the performance of using convolutional downsampling and average pooling is slightly worse than that of using max pooling.

% The loss function makes the network focus on structural information and alleviating the affects of unstable pixels, but may harming to the details. The fusion branch and shortcuts draws the attention to details, but may introduce extra redundancy. Maxpooling is used to lossy and aggressively downsampling. It can reduce redundancy and improve robustness. These components work together and complement each other.
% }

\subsubsection{Fusion method of depth image}
To demonstrate that fusing the raw depth image with feature maps via convolution is better than direct concatenation, we remove the convolution layers used for fusion in the model Ours and name it Ours(concat). The result labeled by ``Experiment on fusion method'' in Table \ref{tab:table5} supports our viewpoint.

\begin{table}[!ht]
\centering
% \captionsetup{font=small, skip=1pt}
\caption{Experiment Result on Weight Modification, Downsampling Implementation and Fusion Method\label{tab:table5}}
\renewcommand{\arraystretch}{0.91}
\resizebox{\linewidth}{!}{%
\begin{tabular}{ccccccc}
\toprule
\multicolumn{1}{c}{Model/Metric} &
  \multicolumn{1}{c}{RMSE} &
  \multicolumn{1}{c}{REL} &
  \multicolumn{1}{c}{MAE} &
  \multicolumn{1}{c}{$\delta$1.05} &
  \multicolumn{1}{c}{$\delta$1.10} &
  $\delta$1.25 \\ \midrule
\multicolumn{7}{c}{Experiment on weight} \\ \midrule
\multicolumn{1}{c}{Baseline} &
  \multicolumn{1}{c}{0.018} &
  \multicolumn{1}{c}{0.027} &
  \multicolumn{1}{c}{0.012} &
  \multicolumn{1}{c}{83.76} &
  \multicolumn{1}{c}{95.67} &
  99.71 \\ 
\multicolumn{1}{c}{Edge Weight Modified} &
  \multicolumn{1}{c}{\textbf{0.017}} &
  \multicolumn{1}{c}{\textbf{0.025}} &
  \multicolumn{1}{c}{\textbf{0.011}} &
  \multicolumn{1}{c}{\textbf{85.34}} &
  \multicolumn{1}{c}{\textbf{96.26}} &
  \textbf{99.75} \\ \midrule
\multicolumn{7}{c}{Experiment on downsampling} \\ \midrule
\multicolumn{1}{c}{Conv Down} &
  \multicolumn{1}{c}{0.016} &
  \multicolumn{1}{c}{0.023} &
  \multicolumn{1}{c}{0.011} &
  \multicolumn{1}{c}{87.16} &
  \multicolumn{1}{c}{96.83} &
  99.80 \\ 
\multicolumn{1}{c}{AvgPooling Down} &
  \multicolumn{1}{c}{0.016} &
  \multicolumn{1}{c}{0.024} &
  \multicolumn{1}{c}{0.011} &
  \multicolumn{1}{c}{87.16} &
  \multicolumn{1}{c}{96.93} &
  99.80 \\ 
\multicolumn{1}{c}{MaxPooling Down} &
  \multicolumn{1}{c}{\textbf{0.015}} &
  \multicolumn{1}{c}{\textbf{0.022}} &
  \multicolumn{1}{c}{\textbf{0.010}} &
  \multicolumn{1}{c}{\textbf{88.18}} &
  \multicolumn{1}{c}{\textbf{97.15}} &
  \textbf{99.81} \\ \midrule
  \multicolumn{7}{c}{Experiment on fusion method} \\ \midrule
  \multicolumn{1}{c}{Ours(concat)} &
  \multicolumn{1}{c}{\textbf{0.015}} &
  \multicolumn{1}{c}{0.023} &
  \multicolumn{1}{c}{0.011} &
  \multicolumn{1}{c}{87.90} &
  \multicolumn{1}{c}{96.68} &
  99.80 \\ 
\multicolumn{1}{c}{Ours} &
  \multicolumn{1}{c}{\textbf{0.015}} &
  \multicolumn{1}{c}{\textbf{0.022}} &
  \multicolumn{1}{c}{\textbf{0.010}} &
  \multicolumn{1}{c}{\textbf{88.18}} &
  \multicolumn{1}{c}{\textbf{97.15}} &
  \textbf{99.81} \\ 
\bottomrule
\end{tabular}%
}
%\vspace{-0.5cm}
\end{table}
\vspace{-0.2cm}

\subsection{Pose Estimation Experiment}
In this experiment, we aim to demonstrate the applicability of our network for downstream tasks and to show that it can improve the accuracy of pose estimate.
To evaluate the performance of pose estimation, we use three evaluation metrics, i.e, the average closest point distance (ADD-S), the area under the ADD-S curve (AUC), and the percentage of ADD-S values that are smaller than 2 \centi\meter.
%\cite{xiang2017posecnn}
% The higher the metrics the stronger the performance.

% This experiment is carried out on the set1 of ClearPose, since Clearpose has an accurate pose annotation without sticker. We use typical network DenseFusion \cite{wang2019densefusion} as pose estimation network. Following the learning strategy of DenseFusion, we train the network on 12G NVIDIA TITAN Xp GPU for 5 epochs with batch size of 128. The margin of refinement is set to 0.03. For fair comparison, we evaluate others works using their released source codes and optimal hyper-parameters or statistics reported in their paper.
Both our method and DFNet are trained on the ClearPose Set 1 and are used to predict the depth of Set 1-Scene 5 for pose estimation purposes. The depth completion result is reported in Table \ref{tab:table6} and a screenshot of the live demonstration is reported in Figure \ref{fig:figure7}. In our experiments, we use DenseFusion \cite{wang2019densefusion}  as the pose estimation method. We trained DenseFusion with the restored depth and tested it on 3,000 randomly selected images. Ideally, a more accurate depth prediction can lead to improved performance in pose estimation. The results of our evaluations, presented in Table \ref{tab:table7}, indicate that the depth restored by our method outperforms DFNet in almost every object in the pose estimation task. This results validate that the depth map given by our method is more appropriate for addressing the downstream task, i.e., pose estimation.
% Depth completion models are trained on ClearPose set 1 and predict the depth of set 1-scene 5 for pose estimation. We train DenseFusion with the restored depth and test on 3k randomly chosen images. Metrics for each object are reported in Table \ref{tab:table7}. Result shows that the depth restored by FDCT outperforms DFNet's in almost every object in pose estimation task.
% \todo{format of tablehead!!}
\begin{table}[!t]
\caption{Depth Completion Results on ClearPose dataset.}
\label{tab:table6}
\centering
\renewcommand{\arraystretch}{0.90}
\begin{tabular}{ccccccc}
\toprule
Model & RMSE           & REL            & MAE            & $\delta$1.05          & $\delta$1.10          & $\delta$1.25          \\ \midrule
DFNet        & 0.048          & 0.038          & 0.033          & 76.36          & 94.22          & \textbf{99.40} \\
Ours         & \textbf{0.045} & \textbf{0.033} & \textbf{0.028} & \textbf{82.15} & \textbf{94.43} & 99.25          \\
\bottomrule
\end{tabular}%
\end{table}

\begin{table}[!t]
\caption{Pose Estimation Results on ClearPose dataset\label{tab:table7}}
\centering
\renewcommand{\arraystretch}{0.95}
\resizebox{\linewidth}{!}{%
\begin{tabular}{ccccccc}
\toprule
Models &
  \multicolumn{3}{c}{DFNet} &
  \multicolumn{3}{c}{Ours} \\ \midrule
Object/Metirc &
  \multicolumn{1}{c}{AUC} &
  \multicolumn{1}{c}{\textless{}2cm} &
  ADD-S(10\%) &
  \multicolumn{1}{c}{AUC} &
  \multicolumn{1}{c}{\textless{}2cm} &
  ADD-S(10\%) \\ 
beaker\_1 &
  \multicolumn{1}{c}{79.07} &
  \multicolumn{1}{c}{\textbf{0.00}} &
  0.68 &
  \multicolumn{1}{c}{\textbf{80.44}} &
  \multicolumn{1}{c}{\textbf{0.00}} &
  \textbf{7.53} \\ 
dropper\_1 &
  \multicolumn{1}{c}{\textbf{67.76}} &
  \multicolumn{1}{c}{61.00} &
  \textbf{48.00} &
  \multicolumn{1}{c}{31.70} &
  \multicolumn{1}{c}{\textbf{65.33}} &
  0.00 \\ 
dropper\_2 &
  \multicolumn{1}{c}{81.09} &
  \multicolumn{1}{c}{\textbf{33.10}} &
  1.78 &
  \multicolumn{1}{c}{\textbf{84.24}} &
  \multicolumn{1}{c}{0.00} &
  \textbf{9.61} \\ 
flask\_1 &
  \multicolumn{1}{c}{84.96} &
  \multicolumn{1}{c}{60.33} &
  42.33 &
  \multicolumn{1}{c}{\textbf{86.71}} &
  \multicolumn{1}{c}{\textbf{68.33}} &
  \textbf{68.00} \\ 
funnel\_1 &
  \multicolumn{1}{c}{78.85} &
  \multicolumn{1}{c}{91.33} &
  0.00 &
  \multicolumn{1}{c}{\textbf{82.91}} &
  \multicolumn{1}{c}{\textbf{98.33}} &
  \textbf{12.33} \\ 
cylinder\_1 &
  \multicolumn{1}{c}{78.77} &
  \multicolumn{1}{c}{48.33} &
  28.67 &
  \multicolumn{1}{c}{\textbf{79.83}} &
  \multicolumn{1}{c}{\textbf{77.00}} &
  \textbf{33.33} \\ 
cylinder\_2 &
  \multicolumn{1}{c}{62.75} &
  \multicolumn{1}{c}{54.67} &
  3.33 &
  \multicolumn{1}{c}{\textbf{75.68}} &
  \multicolumn{1}{c}{\textbf{58.67}} &
  \textbf{29.33} \\ 
pan\_1 &
  \multicolumn{1}{c}{86.76} &
  \multicolumn{1}{c}{13.67} &
  33.33 &
  \multicolumn{1}{c}{\textbf{89.37}} &
  \multicolumn{1}{c}{\textbf{53.67}} &
  \textbf{50.00} \\ 
pan\_2 &
  \multicolumn{1}{c}{88.71} &
  \multicolumn{1}{c}{84.67} &
  44.00 &
  \multicolumn{1}{c}{\textbf{89.73}} &
  \multicolumn{1}{c}{\textbf{90.33}} &
  \textbf{56.00} \\ 
pan\_3 &
  \multicolumn{1}{c}{\textbf{88.90}} &
  \multicolumn{1}{c}{87.67} &
  \textbf{53.33} &
  \multicolumn{1}{c}{88.10} &
  \multicolumn{1}{c}{\textbf{91.00}} &
  48.00 \\ 
bottle\_1 &
  \multicolumn{1}{c}{86.05} &
  \multicolumn{1}{c}{91.53} &
  24.41 &
  \multicolumn{1}{c}{\textbf{88.71}} &
  \multicolumn{1}{c}{\textbf{93.22}} &
  \textbf{31.53} \\ 
bottle\_2 &
  \multicolumn{1}{c}{71.81} &
  \multicolumn{1}{c}{83.16} &
  4.04 &
  \multicolumn{1}{c}{\textbf{77.01}} &
  \multicolumn{1}{c}{\textbf{88.22}} &
  \textbf{13.47} \\ 
stick\_1 &
  \multicolumn{1}{c}{69.53} &
  \multicolumn{1}{c}{32.32} &
  32.66 &
  \multicolumn{1}{c}{\textbf{79.60}} &
  \multicolumn{1}{c}{\textbf{57.58}} &
  \textbf{58.92} \\ 
syringe\_1 &
  \multicolumn{1}{c}{73.03} &
  \multicolumn{1}{c}{31.67} &
  25.67 &
  \multicolumn{1}{c}{\textbf{80.15}} &
  \multicolumn{1}{c}{\textbf{57.00}} &
  \textbf{47.00} \\ 
MEAN &
  \multicolumn{1}{c}{78.43} &
  \multicolumn{1}{c}{55.25} &
  24.45 &
  \multicolumn{1}{c}{\textbf{79.58}} &
  \multicolumn{1}{c}{\textbf{64.19}} &
  \textbf{33.22} \\

  \bottomrule
  \end{tabular}%
}
\vspace{-0.5cm}
\end{table}

\section{Conclusions}
We present FDCT, a novel and efficient end-to-end network for transparent object depth completion. FDCT predicts a rectified depth map using only RGB and depth images as inputs. It outperforms the state-of-the-art methods in terms of accuracy and efficiency, with the smallest number of parameters while running at a fast speed of 70 FPS. Our experiments show that FDCT has strong generalization ability, with excellent results on unseen objects and cross-dataset scenarios. In addition, the improved depth prediction by FDCT can lead to improved performance in pose estimation, demonstrating its potential as a useful auxiliary tool for ordinary pose estimation methods when dealing with transparent objects.
% There is still some works could be done in the future. The hyperparameter $\delta$ in Huber loss is set manually and unchanged. 
Despite the effectiveness, there are still some
works could be done in the future. For example, the hyperparameter $\delta$ in
Huber loss is
set empirically. It is preferred to automate the hyperparameter search.

% In this paper, we propose FDCT, an end-to-end efficient network for transparent objects depth completion. FDCT predicts rectified depth from RGB and depth image only. Our method achieves state-of-the-art performance with least parameters and runs at 70 FPS. Experiment results show that FDCT has a generalization ability to unseen object and cross-dataset. Pose estimation method designed for opaque objects achieves a better performance with depth restored by FDCT, showing its potential to be an auxiliary tool for ordinary pose estimation methods dealing with transparent objects.

\bibliographystyle{IEEEtran}
\bibliography{ref}

\end{document}